\documentclass[lettersize,journal]{IEEEtran}
\usepackage{amsmath,amsfonts}
\usepackage{algorithmic}
\usepackage{algorithm}
\usepackage{array}
\usepackage[caption=false,font=normalsize,labelfont=sf,textfont=sf]{subfig}
\usepackage{textcomp}
\usepackage{stfloats}
\usepackage{url}
\usepackage{verbatim}
\usepackage{graphicx}
\usepackage{cite}
\usepackage{xcolor}
\usepackage{tabularx}
\usepackage{multirow}
\begin{document}

\title{Multi-Modal Graph Neural Network-Inspired Vessel Identification and Tracking for UAV-Based Maritime Surveillance}

\author{Chenjie Zhao,
Jiangtian Nie,
Jiawen Kang,
Yuxu Lu,
Ryan Wen Liu,~\IEEEmembership{Member,~IEEE}

\thanks{Chenjie Zhao and Ryan Wen Liu are with the Sanya Science and Education Innovation Park of Wuhan University of Technology, Sanya 572025, China, with the School of Navigation, Wuhan University of Technology, Wuhan 430063, China. (e-mail: \{cjzhao, wenliu\}@whut.edu.cn) \textit{(Corresponding authors: Ryan Wen Liu.)}}
\thanks{Jiangtian Nie is with the School of Natural and Computing Sciences, University of Aberdeen, Aberdeen AB24 3UE, U.K. (e-mail: jiangtian.nie@abdn.ac.uk)}
\thanks{Jiawen Kang is with the School of Automation, Guangdong University of Technology, Guangzhou 510006, China (emails: kavinkang@gdut.edu.cn)}
\thanks{Yuxu Lu is with the Department of Logistics and Maritime Studies, the Hong Kong Polytechnic University, Hong Kong 999077. (e-mail: yuxulouis.lu@connect.polyu.hk)}
}



\maketitle

\begin{abstract}
    The unmanned aerial vehicle (UAV)-based maritime surveillance is becoming an important part of maritime Internet of Things (IoT) and intelligent transportation systems. It provides a flexible and cost-effective way to large-scale maritime traffic monitoring. However, achieving robust situational awareness remains challenging in practical maritime environments. The main challenges arise from heterogeneous sensing modalities, adverse imaging conditions, and spatial misalignment caused by UAV motion. To address these problems, this paper proposes an airborne multi-modal graph learning-driven framework, termed AMGT, for vessel identification and tracking in UAV-assisted maritime IoT. Specifically, visible and infrared images are first aligned and fused through a multi-parameter perspective transformation model. This improves visual perception under low visibility and complex illumination conditions. Automatic identification system (AIS) messages are then temporally synchronized using interpolation and linear prediction. To detect small and oriented vessels, we develop a mixed-modality MAir-YOLOv8-OBB detector, followed by the MAir-SORT algorithm for stable vessel tracking. We further design a graph learning-driven association network to model temporal dynamics and spatial dependencies among vessels and between vessels and UAVs. A motion-aware coordinate unification (MCU) strategy is also introduced to reduce nonlinear misalignment caused by UAV motion. Experiments on the MAP-M2UAV50 dataset show that AMGT achieves high accuracy and efficiency. The results demonstrate its potential for UAV-assisted maritime IoT applications, including vessel monitoring, traffic situational awareness, and intelligent maritime transportation services. The dataset and code will be available at https://github.com/MAPGroup/AMGT.
\end{abstract}


\section{Introduction}
    \IEEEPARstart{W}{ith} the development of the low-altitude economy, unmanned aerial vehicles (UAVs) are being increasingly integrated with terrestrial, coastal, and shipborne networks \cite{11165289}. This integration forms an air-ground collaborative system and provides new infrastructure for maritime Internet of Things (IoT). In this system, UAVs can collect maritime traffic information from flexible viewpoints. Coastal and shipborne stations can support communication and data transmission. A central controller can further process the collected information and support decision-making. Therefore, UAV-assisted maritime IoT is becoming an important enabling technology for next-generation intelligent transportation systems (ITS), especially for large-scale maritime traffic monitoring and management.

    \begin{figure}[t]
    \centering
    \includegraphics[width=0.5 \textwidth]{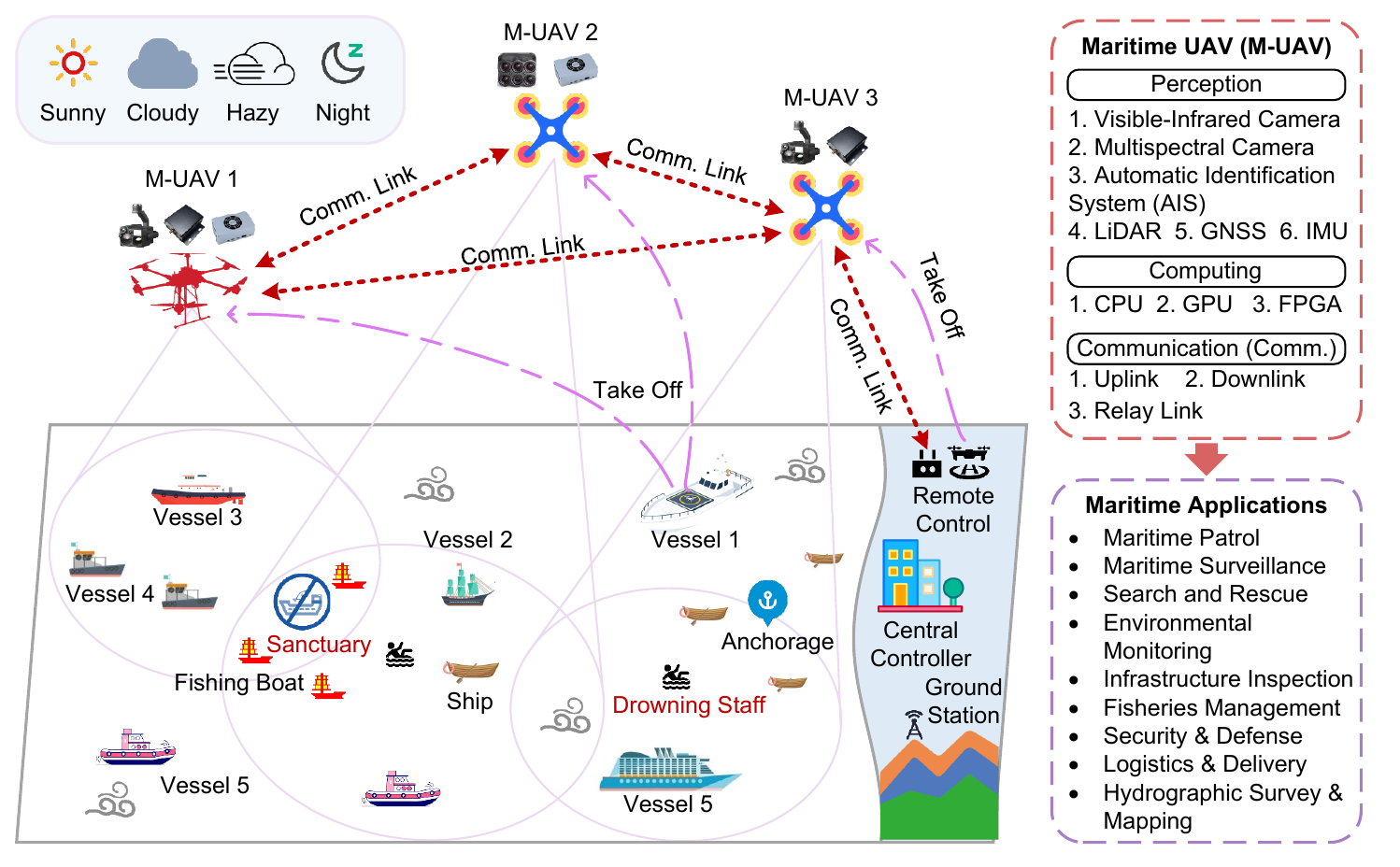}
    \caption{The low-altitude sensing, communication, computing, and control (SCCC) system for UAV-assisted maritime IoT. The system consists of multiple maritime UAVs (M-UAVs), takeoff and landing platforms, ground stations, and a central controller connected by air-ground communication links. With the aid of deep learning-based models and robust algorithms, M-UAVs can support real-time maritime surveillance and collaborative decision-making in complex marine environments.}
    \label{fig: sccc system} 
    % \vspace{-2.0em}
    \end{figure}

    The low-altitude sensing, communication, computing, and control (SCCC) system provides a basic architecture for ITS-oriented maritime surveillance. The SCCC system in Fig. \ref{fig: sccc system} consists of UAV platforms, coastal or shipborne base stations, and a central controller. These components work together to support real-time perception, information exchange, data processing, and decision-making. Different from traditional passive monitoring, maritime surveillance in IoT-enabled ITS is moving toward a closed-loop process. Such a closed-loop process requires continuous traffic sensing, multisource data transmission, vessel-state analysis, and adaptive control. These capabilities are important for trajectory analysis, traffic optimization, collision avoidance, and emergency response \cite{10869306,wang2024deep}. In this way, UAV-assisted SCCC systems can improve maritime traffic safety and efficiency, while also supporting the development of the low-altitude economy.
    Recent studies have explored system-level co-design and optimization for UAV-assisted intelligent systems, particularly the joint coordination of communication, computing, and control resources \cite{10379519,tu2022resource,11142291}. These studies highlight the importance of cross-domain resource coordination for reliable UAV-enabled services.
    
    As illustrated in Fig. 1, reliable maritime situational awareness within such an SCCC system relies on continuous perception and effective processing of heterogeneous information collected from dynamic UAV platforms. Although UAV platforms provide flexible sensing capability, robust maritime situational awareness is still challenging. A single sensing modality is often insufficient for complex maritime environments. Visible cameras provide rich appearance information, but they are sensitive to illumination changes, weather degradation, and limited fields of view. Infrared (IR) cameras can capture object outlines under adverse conditions, such as nighttime, haze, and fog. IR cameras can therefore complement visible-light sensors. However, image-based observations alone cannot reliably provide vessel identity and navigation information such as MMSI, course, and speed. By contrast, the automatic identification system (AIS) provides important maritime information, such as maritime mobile service identity (MMSI), course, and speed. Nevertheless, AIS messages may suffer from low update rates and missing data and may be unavailable for small or non-cooperative vessels. Therefore, it is necessary to fuse visible, IR, AIS, and UAV flight data for accurate and complete maritime perception.

    UAV motion further increases the difficulty of multi-source data fusion \cite{sun2018robust}. Existing vessel association methods \cite{10159572,yang2022multi,qu2023improving,liu2026nav} mainly focus on shore and ship-based surveillance systems. In contrast, UAVs continuously change their altitude and attitude, including pitch, yaw, and roll. These changes introduce nonlinear spatial discrepancies between image coordinates and geographic coordinates. The problem becomes more serious when data from cameras, AIS receivers, and UAV navigation sensors are collected asynchronously. Motion blur, unstable viewpoints, and inaccurate coordinate transformation can also degrade traditional methods based on fixed geometric assumptions. In addition, airborne maritime scenes often contain dense vessel traffic, partial occlusion, adverse weather, sea clutter, and distant small objects. These factors make vessel detection, identification, and tracking more difficult. 
    
    Graph learning provides an effective means of modeling relational information in dynamic maritime scenes. Conventional association methods mainly rely on pairwise geometric distances or trajectory similarities, whose reliability can degrade under asynchronous observations, UAV-induced coordinate variations, and ambiguous vessel motion. In contrast, graph-based modeling represents vessels as nodes and their spatial-temporal relationships as edges, allowing association to exploit not only individual trajectories but also the relative topology among multiple vessels. Such relational structures tend to remain locally consistent even when individual observations are noisy or incomplete. Graph neural networks (GNNs) are therefore well suited to aggregate neighborhood information and learn spatial-temporal dependencies for robust visual–AIS association under dynamic UAV motion.
    
    To address the key challenges of adverse maritime imaging, unstable tracking under dynamic UAV motion, and spatial-temporal misalignment between heterogeneous visual and AIS observations, we propose an airborne multi-modal graph learning-driven framework, termed AMGT, for robust vessel identification and tracking. The main contributions of this work are summarized as follows:
    \begin{itemize}
        \item We introduce MAP-M2UAV50 to address the lack of comprehensive multimodal data for UAV-based maritime surveillance, integrating visible/IR videos, AIS data, UAV flight data, and comprehensive vessel annotations.
        \item We develop a lightweight MAir-YOLOv8-OBB detector to improve small and oriented vessel detection under adverse maritime conditions. It enhances multi-scale feature fusion and uses mixed-modality training to achieve multi-modal perception.
        \item We propose MAir-SORT to improve vessel tracking under false detections, occlusions, and UAV motion. It combines angle-adaptive tracking, camera-motion compensation, and multi-cue association to reduce false tracks and identity switches.
        \item We develop a motion-aware coordinate unification and graph-based association method to address visual–AIS misalignment under dynamic UAV motion, combining motion-compensated spatial alignment with spatio-temporal graph learning for reliable vessel identification.
    \end{itemize}

    The rest of this paper is organized as follows. Section \ref{2} reviews the recent research on object association.
    Section \ref{3} analyzes the self-collected challenging multi-modal dataset. In Section \ref{4}, a multi-modal graph learning-driven framework is proposed and detailed for robust multi-vessel identification and tracking. Section \ref{5} conducts comprehensive experiments to demonstrate the superiority and efficiency of our method. Finally, Section \ref{6} summarizes the main contributions and future perspectives of this work.

\section{Related Work} 
\label{2}
    UAV-based maritime surveillance involves locating objects in aerial images, maintaining their identities across frames, and linking visual observations with AIS information. Related studies are reviewed according to these three tasks, namely object detection, object tracking, and visual--AIS vessel association.

    \subsection{UAV-Based Maritime Object Detection}
    Airborne maritime detection must localize small objects across changing scales, orientations, and imaging conditions. Existing approaches include two-stage detectors and single-stage models from the YOLO family. Zhao and Song \cite{zhao2024anchor} optimized anchor boxes for two-stage detectors using the object-size characteristics of SeaDronesSee, improving the localization of difficult maritime search-and-rescue targets. For oriented detection, YOLOv8-OBB and YOLO11-OBB \cite{Jocher_Ultralytics_YOLO_2023} predict rotated boxes that describe object position, size, and orientation. This representation retains geometric information that is not explicitly encoded by horizontal boxes.

    In addition to model design, recent research considers the availability and diversity of maritime training data. Zhao et al. \cite{zhao2025construction} constructed a virtual maritime search-and-rescue dataset with varied weather, lighting, waves, and UAV shooting modes. Such work complements detector adaptation by expanding the observation conditions represented in training data. Nevertheless, distant targets may occupy only a small number of pixels, while changing viewpoints and illumination affect their apparent shape and contrast. Reliable localization across these conditions remains a central difficulty in airborne maritime detection.

    \subsection{UAV-Based Maritime Object Tracking}
    Tracking-by-detection maintains object identities by associating observations across frames. SORT \cite{7533003} combines Kalman prediction with Hungarian assignment, providing an efficient motion-based approach. However, discarding low-confidence detections can interrupt otherwise valid tracks. ByteTrack \cite{zhang2022bytetrack} addresses this issue by additionally associating low-confidence detections with existing tracklets. BoT-SORT \cite{aharon2022bot} further combines motion and appearance information with camera-motion compensation. These approaches address different sources of tracking error, including uncertain detections and changes in the camera viewpoint.

    For UAV-based maritime tracking, Yang et al. \cite{yang2024seayoulater} introduced metadata-guided re-identification using UAV position, altitude, and camera orientation to reconnect short tracklets and maintain long-term identities. This approach supplements image observations with flight information, whereas image-based compensation estimates camera motion from visual correspondences. Maritime tracking nevertheless remains difficult when small vessels have similar appearances and observations are interrupted by occlusion or abrupt camera movement. In these situations, maintaining identity continuity requires handling both vessel motion and the apparent image motion induced by the UAV.

    \subsection{Visual-AIS Vessel Association}
    Non-learning-based association methods typically rely on geometric constraints, trajectory similarity, and optimization-based matching. Geometric approaches associate heterogeneous observations according to position, field of view, course, or other physically interpretable constraints \cite{9594203} \cite{10833660}, while trajectory-based methods exploit temporal motion consistency using predefined distance measures. For example, Guo et al. \cite{10159572} employed E-FastDTW for asynchronous trajectory matching in inland vessel surveillance. These methods are generally interpretable and do not require training data. However, their performance depends strongly on accurate coordinate alignment, temporal synchronization, and predefined similarity metrics, which can become unreliable under dynamic UAV motion and ambiguous vessel trajectories.

    Learning-based approaches provide an alternative to predefined trajectory distances. Recurrent architectures, including LSTM and GRU \cite{graves2012long} \cite{chung2014empirical}, model sequential dependencies, while GCN \cite{kipf2016semi} and GAT \cite{velickovic2017graph} aggregate information over graph neighborhoods. In maritime multisensor association, Yang et al. \cite{yang2022multi} combined graph representations, self-cross attention, and optimal transport to associate AIS and radar tracks. The method uses relational information to address dense and asynchronous observations. However, its AIS--radar setting differs from visual--AIS association under oblique aerial views. Changes in UAV position and attitude alter the mapping between image and geographic coordinates, while missing observations introduce further uncertainty. Learned representations therefore need to be considered together with coordinate handling and temporal alignment when assessing association reliability.

\section{Multi-Modal Maritime Aerial Dataset}
\label{3}
    \begin{table*}[]
    \centering
    \caption{The UAV aerial datasets in maritime scenarios. Our dataset comprises data from four distinct modalities. The rotated bounding boxes (RBboxes) were manually annotated. Additionally, various weather conditions, including sunny, hazy, night, and sun glare scenarios, are encompassed.}
    \begin{tabular}{cccccccccc}
    \hline
    \textbf{\begin{tabular}[c]{@{}c@{}}UAV   Aerial \\ Dataset\end{tabular}} & \textbf{\begin{tabular}[c]{@{}c@{}}Maritime  \\ Scenario\end{tabular}}    & \textbf{Perception Task}                                                                & \textbf{Images} & \textbf{Videos} & \textbf{Data Modal}                                                   & \textbf{\begin{tabular}[c]{@{}c@{}}Total \\ Anno.\end{tabular}} & \textbf{\begin{tabular}[c]{@{}c@{}}Num. of \\ Class\end{tabular}} & \textbf{\begin{tabular}[c]{@{}c@{}}Altitude\\ (m)\end{tabular}} & \textbf{\begin{tabular}[c]{@{}c@{}}Oriented \\ RBbox\end{tabular}} \\ \hline
    Seagull \cite{8115215}                                                                 & \begin{tabular}[c]{@{}c@{}}Sea Monitoring \\ \& Surveillance\end{tabular} & \begin{tabular}[c]{@{}c@{}}Object Detection \\ \& Tracking\end{tabular}                 & ——              & 19              & \begin{tabular}[c]{@{}c@{}}Visible\\ Infrared\end{tabular}            & ——                                                              & 6                                                                 & 150-300                                                         & $\times$                                                                 \\
    AFO \cite{gasienica2021ensemble}                                                                      & Search \& Rescue                                                          & Object Detection                                                                        & 3640            & 50              & Visible                                                               & 12K                                                             & 9                                                                 & 30-80                                                           & $\times$                                                                 \\
    MOBDrone \cite{cafarelli2022mobdrone}                                                                & Search \& Rescue                                                          & Object Detection                                                                        & 125K            & 66              & Visible                                                               & 180K                                                            & 5                                                                 & 10-60                                                           & $\times$                                                                 \\
    Zhao et al.  \cite{zhao2023multiship}                                                            & \begin{tabular}[c]{@{}c@{}}Maritime   \\ Surveillance\end{tabular}        & Object Detection                                                                        & 18K             & 3               & Visible                                                               & ——                                                              & 1                                                                 & 120                                                             & $\times$                                                                 \\
    MAP-M2UAV50                                                                     & \begin{tabular}[c]{@{}c@{}}Maritime  \\  Surveillance\end{tabular}        & \begin{tabular}[c]{@{}c@{}}Object Detection, \\ Tracking \\ \& Association\end{tabular} & 12K             & 50              & \begin{tabular}[c]{@{}c@{}}Visible \\ Infrared \\  AIS \& \\ Flight data \end{tabular} & 70K                                                             & 3                                                                 & 150-300                                                         &  \checkmark                                                                 \\ \hline
    \end{tabular}
    \label{tab: dataset}
    \end{table*}

    A challenging maritime aerial multimodal dataset, termed MAP-M2UAV50, is proposed and contributed for UAV-based maritime surveillance. It integrates various heterogeneous data sources collected from the UAV-based platform, specifically including:
    \begin{itemize}
    \item Visible images/videos.
    \item IR images/videos.
    \item UAV flight data, i.e., internal sensing data, such as GNSS (longitude, latitude, and height) and IMU (including 3-axis gyroscope, accelerometer, and magnetometer) data, etc.
    \item AIS data (MMSI, position, speed, course, and other vessel information).
    \end{itemize}
    
    A comprehensive summary of the dataset specifications, including data scale, modalities, flight altitude, etc., is provided in Table \ref{tab: dataset}. The 
    manually annotated labels consist of rotated bounding boxes, tracking ID, and corresponding MMSI of vessels. Distinct from existing shore-based and vessel-borne benchmarks, e.g., FVessel \cite{10159572} and WaterScenes \cite{10571852}, as well as UAV-based tracking datasets in general scenarios \cite{kiefer20231st}, MAP-M2UAV50 introduces unprecedented challenges and superior research value, characterized by following aspects:
    
    \subsubsection{Image degradation under adverse weather conditions}
    Poor visibility due to adverse weather conditions, e.g., hazy and night days, always causes severe degradation in visible image data, undermining the robustness and accuracy of airborne object detection and tracking.

    \subsubsection{Multi-scale and sparse feature in airborne images}
    Due to the broad surveillance coverage of UAVs, distant objects appear as small objects with sparse pixel features in airborne images. Moreover, the continuous variations in object scale also decrease rotated object detection accuracy.
    
    \subsubsection{Motion blur caused by the UAV-based dynamic platform}
    The rapid movement and dynamic attitude changes, e.g., yaw and pitch, of the UAV and its onboard camera always lead to image blur and jittering, making it difficult to detect and tracking steadily.  

    \subsubsection{Spatio-temporal misalignment for heterogeneous data}
    A critical technical challenge arises from the asynchronous and spatially discrepant nature of UAV attitude, visual observations, and AIS data during dynamic flight. Achieving high-precision visual-AIS object association is fundamental to enabling robust maritime multimodal airborne data fusion.

\section{The proposed AMGT framework} 
\label{4}
\begin{figure*}
    \centering
    \includegraphics[width=1 \textwidth]{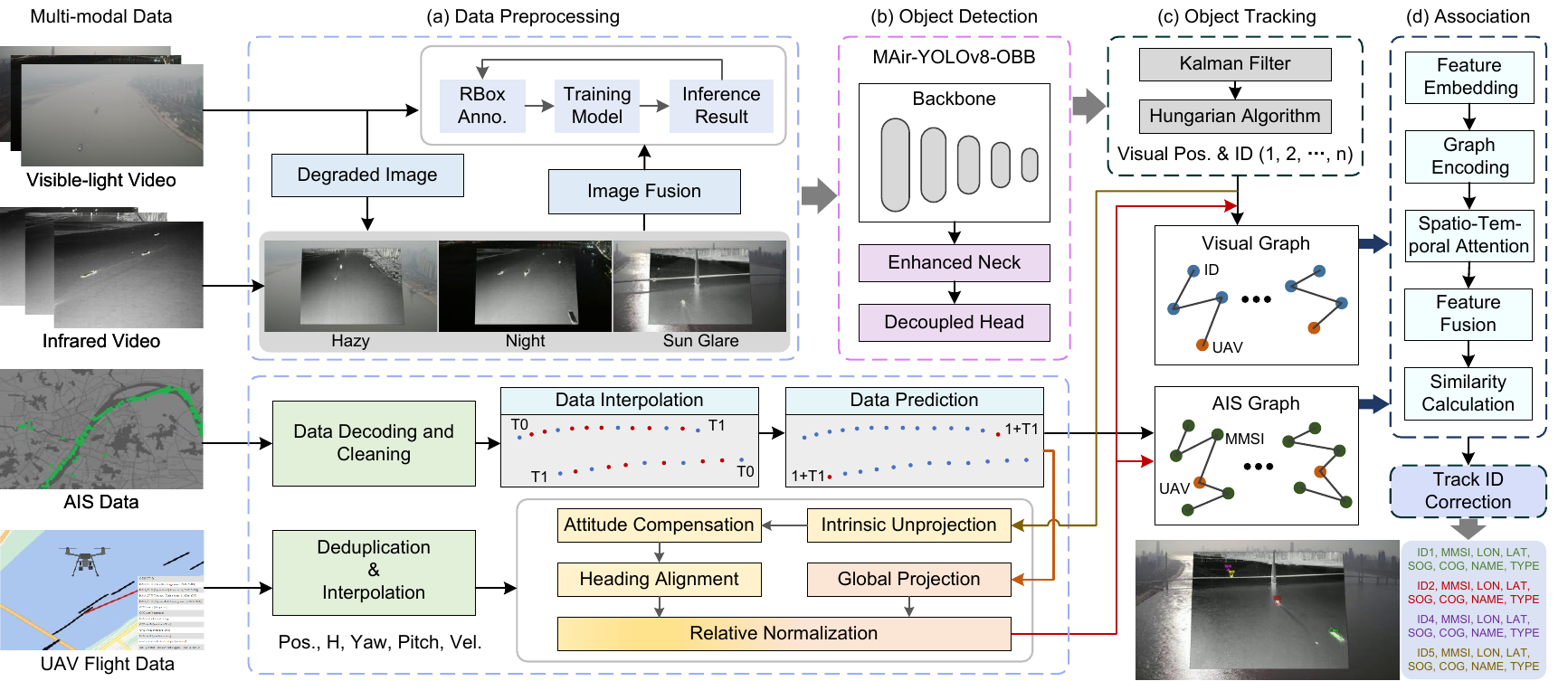}
    \caption{The overall architecture of our AMGT, a multi-modal graph learning-driven multi-vessel identification and tracking framework. Four data modalities are input, i.e., visible and IR videos, AIS, and UAV flight data. The data preprocessing mainly includes image fusion under adverse weather, as well as AIS decoding, cleaning, and prediction. Detection is performed by an enhanced MAir‑YOLOv8‑OBB network, followed by MAir-SORT vessel tracking method. Visual and AIS positions are spatially aligned and normalized to form two graphs, which are then associated using a spatio‑temporal graph neural network. Tracking IDs are ultimately corrected to output identified and tracked vessels.}
    \label{fig: method} 
    \vspace{-1.0em}
    \end{figure*}

    In this section, we detail the proposed airborne multi-modal graph learning-driven framework, i.e., AMGT shown in Fig. \ref{fig: method}, for robust multi-vessel identification and tracking in the UAV-assisted SCCC system illustrated in Fig. \ref{fig: sccc system}. It consists of four modules, i.e., multimodal data preprocessing, MAir-YOLOv8-OBB detection network, MAir-SORT vessel tracking, and graph learning-inspired association network. These modules form a sequential and tightly coupled pipeline, with each module addressing a specific challenge arising from the dynamic UAV-based multimodal sensing platform.

    \subsection{Multimodal Data Preprocessing}
        \subsubsection{Visual Data}
            Single-modality visual sensing is insufficient for UAV-based maritime surveillance under all-weather and all-lighting conditions. Visible images often suffer from severe quality degradation at night or in foggy weather. During dynamic flights, UAVs may also encounter rapid illumination changes, motion blur, and water-background noise, making distant small vessels difficult to detect. IR sensors can capture thermal radiation differences when visible images degrade. Therefore, visible--IR fusion provides complementary information and improves perception robustness in complex maritime environments.

            Due to differences in sensing hardware, visible and IR cameras usually have different FOV, sampling rates, and mounting positions. Lens distortion and physical offsets may further cause spatial misalignment. To address these issues, we adopt a frame-rate synchronization strategy based on linear interpolation. Let $I_{\text{IR}}^{(k)}$ and $I_{\text{IR}}^{(k+1)}$ denote two consecutive IR frames sampled at times $t_k$ and $t_{k+1}$. For a visible frame sampled at time $t_{vis}$ where $t_k \leq t_{vis} \leq t_{k+1}$, the synchronized IR frame $\hat{I}_{ir}$ is synthesized as from
            \begin{equation}
            \hat{I}_{\text{IR}} = (1-\alpha) I_{\text{IR}}^{(k)} + \alpha I_{\text{IR}}^{(k+1)},
            \end{equation}
            with $\alpha = \frac{t_{vis} - t_k}{t_{k+1} - t_k}$. A multi-parameter perspective transformation model is then proposed for visible-IR alignment. The model maps synchronized IR images into the visible coordinate system. 
            A homography transformation matrix $\mathbf{H} \in \mathbb{R}^{3 \times 3}$ is constructed as follows,
            
            \begin{itemize}
            	    \item Define transformation parameters. To model the physical displacement and perspective distortion between visible and IR sensors, we define a parameter set $\Theta = \{t_x, t_y, s, \sigma_{far}, \mathbf{c}_{offset}\}$. Here, $(t_x, t_y)$ denotes the translation vector, $s$ represents the global scaling factor, and $\mathbf{c}_{offset}$ accounts for the deviation of the principal point. Specifically, $\sigma_{far}$ is introduced as a perspective-scaling factor to correct the trapezoidal distortion caused by camera tilt angle.
            	
            	    \item Generate correspondence points. We utilize four vertex coordinates of the source IR image $\mathbf{P}_{src}$ as control points. The corresponding target control points $\mathbf{P}_{dst}$ in the visible coordinate system are analytically derived by applying a parameterized projection function $f(\mathbf{P}_{src}; \Theta)$. It explicitly adjusts the coordinates of the upper vertices relative to the projection center based on $\sigma_{far}$ to simulate depth-induced deformation, followed by global affine transformations.
            	
            	    \item Construct and apply the transformation matrix. By solving the linear system $\mathbf{P}_{dst} \sim \mathbf{H} \mathbf{P}_{src}$ using the direct linear transform algorithm, we obtain the homography transformation matrix $\mathbf{H}$. The synchronized IR image is then warped via $\mathbf{H}$ to achieve geometric alignment with the visible image. The final registered output is obtained by region-selective fusion, i.e.,
            	
            	    \begin{equation}
            		    \label{eq:fusion}
            		    I_{\text{fused}} = \mathcal{W}(\hat{I}_{\text{IR}}, \mathbf{H}) \odot \mathbf{M} + I_{\text{VIS}} \odot (\mathbf{1} - \mathbf{M}),
            		    \end{equation}
            	
            	    \noindent where $\mathcal{W}(\cdot)$ denotes the perspective warping operation, $\mathbf{M}$ is the binary validity mask of the warped IR region, which is smoothed by a Gaussian kernel at the boundaries. $\odot$ represents the element-wise multiplication.
            	\end{itemize}

            To further improve fusion quality, a spatial superposition mechanism is introduced to construct an IR region mask, where IR regions provide salient thermal cues and the remaining regions preserve visible appearance information.

            \begin{figure}
            \centering
            \includegraphics[width=0.5 \textwidth]{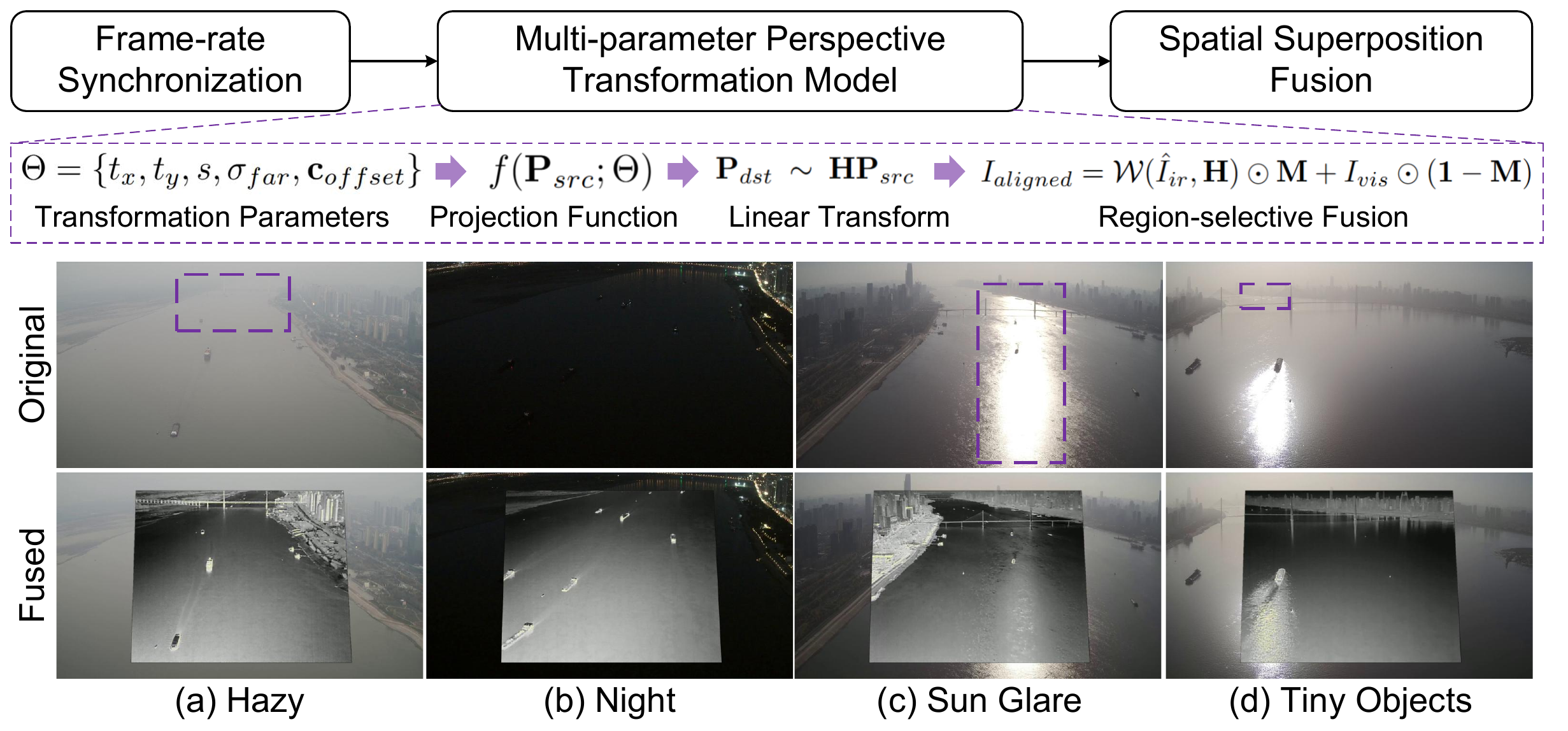}
            \caption{From top to bottom: flow diagram of visible and IR image fusion method, the input original low visibility images, and the output clear fused images. Four challenges are presented: (a) Hazy, (b) Night, (c)Sun Glare, (d) Tiny Objects.}
            \label{fig: image fusion} 
            % \vspace{-2.0em}
            \end{figure}

            For detection and tracking, we sample one frame per second from both visible and IR videos. Since many surface objects show slow pixel-level motion in consecutive aerial frames, images for detector training are sampled every 1 to 10 seconds to reduce redundant visual scenes and alleviate overfitting. Notably, AMGT mainly exploits relative spatial relationships among vessels and between vessels and the UAV. Therefore, it does not rely on highly accurate absolute camera calibration. 

        \subsubsection{AIS Data}
            AIS data processing includes decoding, cleaning, interpolation, and prediction. The data decoding and cleaning are described in detail in our previous work \cite{10833660}. To achieve time synchronization of multimodal data, AIS data is linearly interpolated to obtain measurements at one-second intervals across the entire acquisition period. Subsequently, interpolated AIS trajectories (MMSI) are manually associated with tracked trajectories (track ID) \footnote{We exploit ArcGIS software to project AIS and UAV positions and determine true AIS data of vessels in UAV aerial videos.}. It is regarded as multi-modal ground-truth for evaluating the accuracy of asynchronous data association.

            To achieve real-time association between visual and AIS objects, our method leverages historical AIS data to forecast it at future time steps. Due to limited onboard computing on UAVs, we employ a simple linear regression model for AIS prediction. The predicted AIS data are then used for heterogeneous data association and vessel identification. When new AIS data arrive, we switch to the real AIS data and re-interpolate historical AIS data to refresh the prediction with interpolated values. Moreover, the combination of the simple linear regression model and dynamic update effectively prevents the accumulation of AIS prediction errors.

        \subsubsection{UAV Flight Data}
            We extract timestamps, position, altitude, yaw, pitch, and roll angles of the UAV and camera from UAV flight data at 1Hz sample rate. The flight data analysis can be found in Fig. \ref{fig: coor_unification}. The UAV position and velocity vary over time, which further increases the difficulty of maritime object tracking and heterogeneous data association.

    \subsection{Maritime UAV Object Detection}
            Building upon the experimental analysis from our previous work \cite{10833660}, the YOLOv8-OBB \cite{Jocher_Ultralytics_YOLO_2023} architecture demonstrates superior performance in UAV-based angle-adaptive vessel detection. In this work, an enhanced YOLOv8-OBB network (termed MAir-YOLOv8-OBB) is specifically designed to address the challenges of adverse weather and distant small object detection.

            \subsubsection{Training Strategy}
            
            In maritime environments, visible imagery is highly susceptible to degradation cased by haze, low light, and sun glare conditions, rendering distant small objects nearly indistinguishable. To mitigate the problem, IR imagery is introduced to leverage robust thermal feature capabilities. Based on extensive experimental analysis, we adopt a unified mixed-modality training strategy that jointly trains on visible and infrared images. It is theoretically grounded in the fact that fused imagery contains dual-domain features. The model must simultaneously learn to extract fine-grained visual details and IR thermal features. By exposing the model to a mixed distribution of modalities during training, its generalization capability and detection robustness can be significantly enhanced across diverse environmental conditions.
            
            \subsubsection{Network Structure}
            
\begin{figure*}
            \centering
            \includegraphics[width=1 \textwidth]{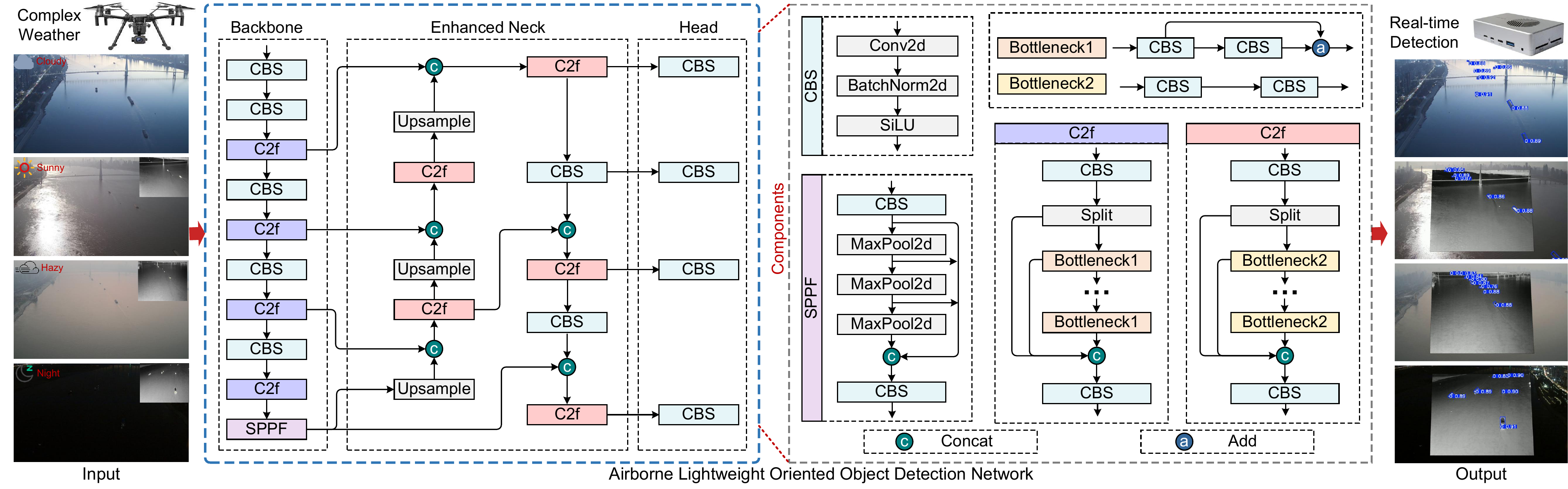}
            \caption{The diagram of our MAir-YOLOv8-OBB detection network. The input comprises UAV aerial imagery across varying weather and water conditions. Image features are extracted and fused via the backbone and enhanced neck, followed by the prediction of rotated bounding boxes through the detection head.}
            \label{fig: yolov8-obb} 
            \vspace{-1.0em}
            \end{figure*}

            To enhance detection accuracy for distant small objects, an additional high-resolution detection head is incorporated into YOLOv8-OBB network. It is designed to capture fine-grained spatial features, significantly boosting the recall rate for small objects. The proposed MAir-YOLOv8-OBB network is illustrated in Fig. \ref{fig: yolov8-obb}. The backbone network is primarily composed of CBS, C2f, and SPPF modules. The C2f module is used for multi-scale feature extraction, and SPPF captures these features through pooling windows of different sizes. The neck employs a path aggregation network structure to perform multi-scale feature fusion, concatenating feature maps from four distinct scales via upsampling operations. The head adopts a decoupled design to independently regress feature maps for classification, bounding box coordinates, and orientation angles.

            \subsubsection{Loss Function}

            The loss function of the proposed MAir-YOLOv8-OBB is defined as a weighted sum of three components, i.e., box regression loss $\mathcal{L}_{box}$, classification loss $\mathcal{L}_{cls}$, and distribution focal loss $\mathcal{L}_{dfl}$. The total loss for rotated object detection is expressed as follows

            \begin{equation}
                \mathcal{L}_{det} = \lambda_{box}\mathcal{L}_{box} + \lambda_{cls}\mathcal{L}_{cls} + \lambda_{dfl}\mathcal{L}_{dfl},
            \end{equation}

            \noindent where $\lambda_{box}$, $\lambda_{cls}$, and $\lambda_{dfl}$ are their weights of respective loss components with values of 7.5, 0.5, and 1.5, respectively. 
            
            Specifically, the classification loss $\mathcal{L}_{cls}$ employs binary cross entropy (BCE) to measure the discrepancy between predicted class scores and target scores. For the rotated bounding box loss $\mathcal{L}_{box}$, unlike horizontal detection methods that use CIoU, ProbIoU loss \cite{10382963} is utilized. Since standard IoU is non-differentiable for rotation, ProbIoU models oriented bounding boxes as 2D Gaussian distributions and computes the similarity based on the Hellinger distance. It ensures robust optimization for arbitrary orientations and avoids boundary discontinuities. The loss is computed as follows

            \begin{equation}
                \mathcal{L}_{box} = \frac{1}{N_{pos}} \sum_{i \in \mathcal{P}} w_i \left( 1 - \text{ProbIoU}(\mathbf{B}_{pred}^{(i)}, \mathbf{B}_{gt}^{(i)}) \right),
            \end{equation}

            \noindent where $\mathbf{B}_{pred}$ and $\mathbf{B}_{gt}$ are the predicted and ground-truth rotated boxes, $w_i$ is the weight derived from target scores, and $N_{pos}$ is the normalization factor. Additionally, $\lambda_{dfl}$ is integrated to refine the regression of bounding box boundaries. It models the distances from anchor points to box edges as a general distribution, enforcing the network to focus on values near the label, which is particularly effective for learning the spatial uncertainty of vessel boundaries.

    \subsection{Angle-Adaptive Vessel Tracking}
        In this section, an angle-adaptive SORT-based algorithm, termed MAir-SORT, is proposed to achieve stable vessel tracking under complex conditions. which integrates motion modeling with false alarm suppression and ID stabilization strategies. It handles the periodicity of box orientation, compensates for image motion caused by the UAV, combines geometric and motion cues during association, and recovers temporarily lost tracks through a state-priority cascade.

        \subsubsection{Oriented State Modeling and Camera-Motion Compensation}

        Following our previous work \cite{10833660}, each rotated track is represented by the nine-dimensional Kalman state
        
        \begin{equation}
            \mathbf{x} = [u, v, s, r, \theta, \dot{u}, \dot{v}, \dot{s}, \dot{r}]^T,
        \end{equation}

        \noindent where $(u,v)$ is the box center, $s$ is its area, $r$ is the aspect ratio, and $\theta$ is the orientation angle. The corresponding observation is $\mathbf{z}=[u,v,s,r,\theta]^T$. Since a rotated rectangle is unchanged after a $180^{\circ}$ rotation, the angular innovation is computed as

        \begin{equation}
            \Delta\theta = \operatorname{mod}(\theta_{d}-\theta_{p}+90^{\circ},180^{\circ})-90^{\circ},
        \end{equation}
        
        \noindent where $\theta_{d}$ and $\theta_{p}$ denote the detected and predicted orientations. This operation prevents discontinuous updates near the angular boundary. Before data association, sparse optical-flow correspondences estimate an affine warp between adjacent frames. The warp updates the predicted center, translational velocity, box area, orientation, covariance, and last observed box. This compensation reduces the apparent image displacement caused by UAV motion before vessel motion is predicted and associated.

        \subsubsection{Multi-Cue Association}

        The rotated-box SORT baseline computes pairwise rotated IoU and applies Hungarian assignment. IoU alone can be unreliable for small oriented vessels because a small center displacement may cause a large reduction in overlap. MAir-SORT therefore constructs the association cost

        \begin{equation}
        \begin{split}
            C_{ij} ={}& 0.55(1-\operatorname{IoU}^{rot}_{ij})
            +0.30\min(d^{c}_{ij}/g_j,1) \\
            &+0.15\min(d^{s}_{ij}/2,1)
            +0.20d^{dir}_{ij}+0.10d^{\theta}_{ij},
        \end{split}
        \end{equation}
        
        \noindent where $d^{c}_{ij}$ is the center distance normalized by the larger box diagonal, $d^{s}_{ij}$ is the absolute logarithmic area ratio, $d^{dir}_{ij}$ measures the difference between the predicted motion and observed displacement, and $d^{\theta}_{ij}=|\Delta\theta|/90^{\circ}$. The gate $g_j$ is tighter for active tracks and wider for lost tracks. Pairs outside the center-distance and size gates are rejected before Hungarian assignment. A feasible pair is accepted when its rotated IoU exceeds 0.05 or when its normalized center distance and area difference satisfy the corresponding acceptance gates.

        \subsubsection{State-Priority Cascade and Lost-Track Recovery}

        Tracks are divided into confirmed, lost, and tentative groups. Unmatched detections are associated with these groups in that order, which gives established trajectories priority over uncertain tracks. A new track becomes confirmed after three successful updates. A lost track remains eligible for matching for up to 40 frames and uses wider center-distance gates than an active track. This design allows the same internal identity to be recovered after a short detection interruption. Only tracks matched to a detection in the current frame are reported, and the reported rotated box uses the latest observed geometry.

    \subsection{Motion-Aware Coordinate Unification}
        To construct a heterogeneous graph in which visual and AIS objects share a spatial reference, image-plane coordinates and geodetic coordinates should be mapped into a common metric space. We establish a local East–North–Up (ENU) coordinate system centered at the UAV’s initial position. The dynamic UAV and camera attitudes can shift the apparent positions of visual objects, making a simple linear mapping insufficient. Therefore, we introduce a motion-aware coordinate unification (MCU) module to compensate for UAV motion and estimate the positions of visual objects in the common reference frame.

        \begin{figure}
        \centering
        \includegraphics[width=0.40 \textwidth]{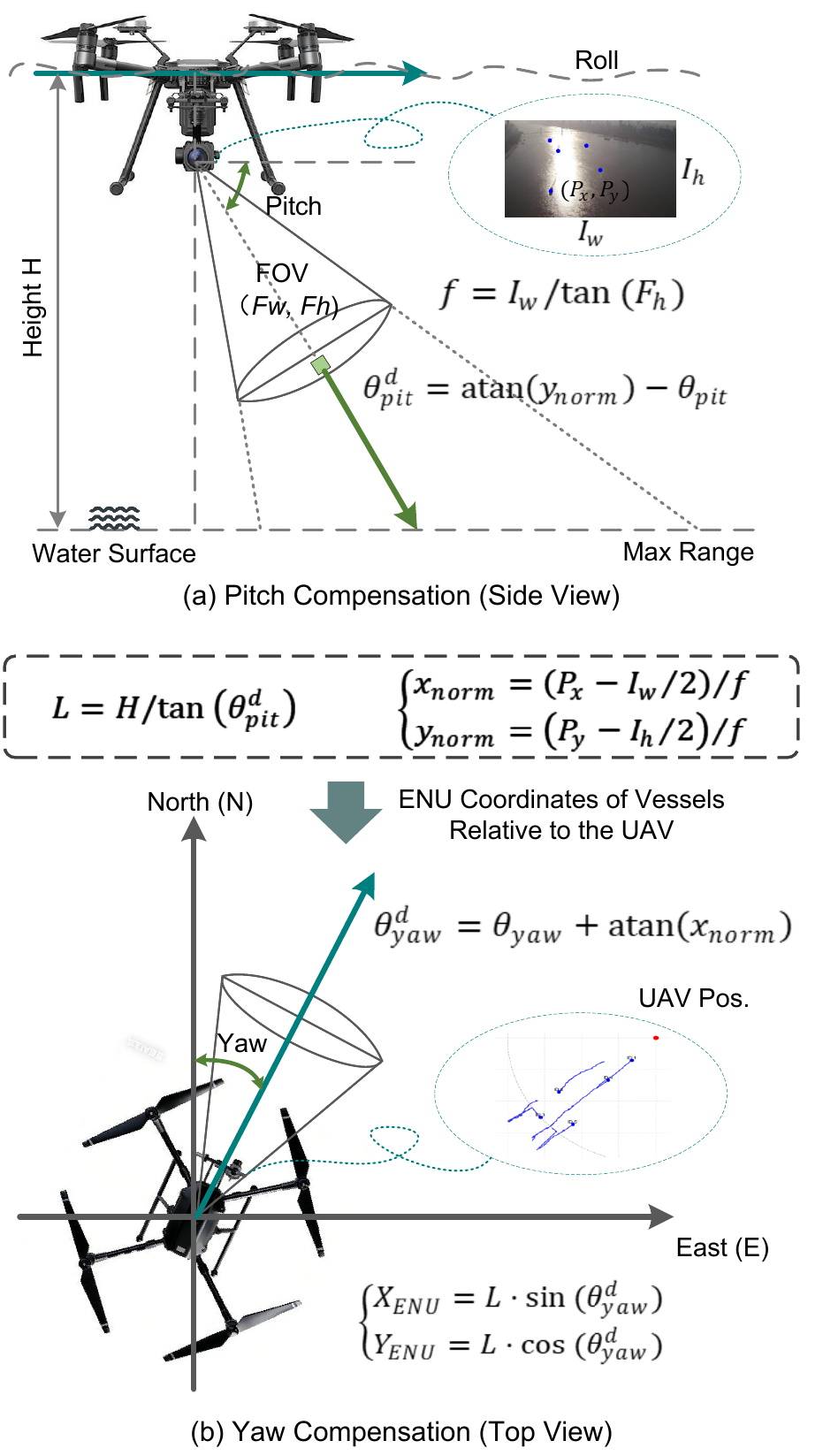}
        \caption{The illustration of attitude compensation for UAV motion. (a) Pitch compensation in the side view. (b) Yaw compensation and relative ENU projection in the top view.}
        \label{fig: compensation} 
        % \vspace{-2.0em}
        \end{figure}

        Formally, let $\mathbf{p} = [P_x, P_y]^T$ denote the pixel coordinate of a detected bounding box center in the image plane. The transformation chain to the unified world coordinate $\mathbf{P}_{w} = [X_w, Y_w]^T$ consists of three sequential compensation steps.

        Step 1: Intrinsic Back-Projection. First, we convert the pixel coordinate $\mathbf{p}$ into the normalized camera coordinate $\mathbf{c} = [x_{norm}, y_{norm}, 1]^T$, i.e.,

        \begin{equation}
        \left\{\begin{array}{c}
            x_{norm} = (P_x - I_w/2) / f, \\
            y_{norm} = (P_y - I_h/2) / f,
        \end{array} \right.
        \end{equation}

        \noindent where $I_w$ and $I_h$ are the image width and height, respectively, and $f=I_w/\tan(F_h)$ is the focal length derived from the horizontal field of view $F_h$.
        
        Step 2: Attitude Compensation (i.e., pitch and yaw). As shown in Fig. \ref{fig: compensation}, the core challenge lies in rectifying the perspective distortion caused by the UAV's attitude. We define the UAV's pitch angle as $\theta_{pit}$ (negative for look-down) and yaw angle as $\theta_{yaw}$ (0 for North, clockwise). Unlike standard rotation matrices, we employ a geometric decomposition approach to ensure numerical stability. The compensated depression angle $\theta_{pit}^{d}$ is obtained from the normalized vertical coordinate and the UAV's pitch, i.e.,

        \begin{equation}
        \theta_{pit}^{d} = \operatorname{atan}(y_{norm}) - \theta_{pit}.
        \end{equation}

        The horizontal ground distance $L$ from the UAV to the object is then calculated as $L = H / \tan(\theta_{pit}^{d})$, where $H$ is the UAV's relative altitude. Subsequently, the compensated heading $\theta_{yaw}^{d}$ is obtained from the normalized horizontal coordinate and the UAV's yaw, i.e.,

        \begin{equation}
        \theta_{yaw}^{d} = \theta_{yaw} + \operatorname{atan}(x_{norm}).
        \end{equation}

        The camera's ego-motion is effectively decoupled, and the object's relative ENU coordinates are calculated as $[X_{ENU},Y_{ENU}]^T=[L\sin(\theta_{yaw}^{d}),L\cos(\theta_{yaw}^{d})]^T$.
        
        Step 3: Global Translation Alignment. The UAV's GPS displacement is compensated to align with the AIS data. Let $\mathbf{P}_{uav}(t)$ be the UAV's position at time $t$ projected into the local ENU frame. The absolute coordinate of the visual object is given by

        \begin{equation}
        \mathbf{P}_{w}^{vis} = \mathbf{P}_{uav}(t) + [X_{ENU},Y_{ENU}]^T.
        \label{eq_L(i, j)}
        \end{equation}

\begin{figure*}
        \centering
        \includegraphics[width=1 \textwidth]{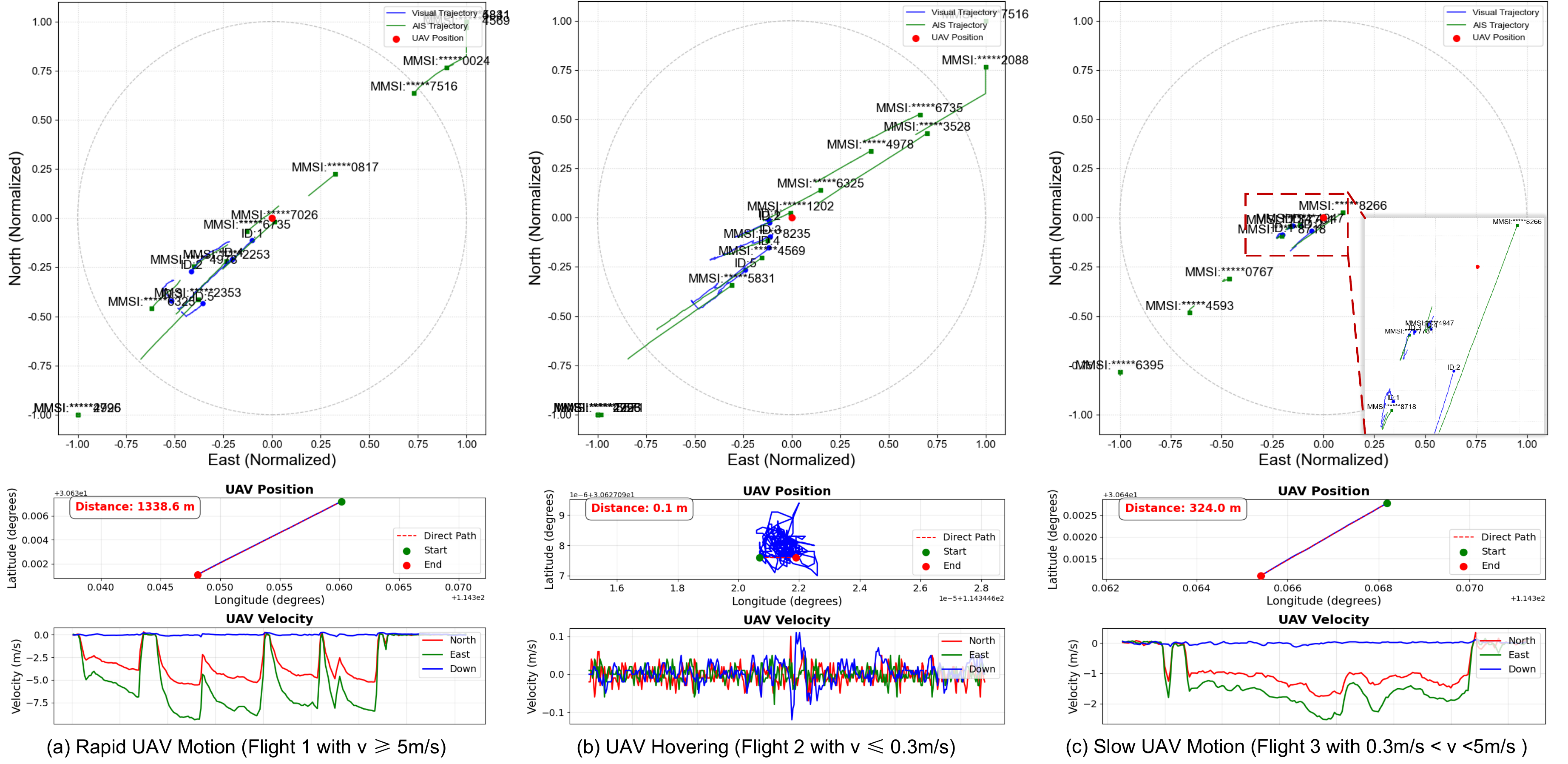}
        \caption{The motion-aware coordinate unification (MCU) across three UAV flight scenarios: (a) Rapid Motion (Flight 1), (b) Hovering (Flight 2), and (c) Slow Motion (Flight 3). The top displays the trajectories of visual and AIS objects projected onto the motion-compensated metric plane, and the bottom depicts UAV position and velocity variations. The rough alignment of trajectories in all cases demonstrates that the unification method is robust and insensitive to complex maneuvers of the UAV.} 
        \label{fig: coor_unification} 
        \vspace{-1.0em}
        \end{figure*}

        For AIS data, the geodetic coordinates $(\lambda, \varphi)$ are directly projected into the same local ENU space to obtain $\mathbf{P}_{w}^{ais}$. By unifying visual and AIS objects into the motion-compensated metric space (shown in Fig. \ref{fig: coor_unification}), the subsequent graph learning module can explicitly model spatial interactions between visual and AIS graph nodes, which is robust to motions of the UAV and camera.

    \subsection{Graph Learning-Inspired Association}  
    \label{sec: association}
        To robustly associate heterogeneous AIS and visual trajectories on the highly dynamic flight platform, we improve a dual-stream graph learning network inspired by \cite{yang2022multi}. The network treats vessel trajectories as inputs and learns a similarity metric by hierarchically capturing local temporal dynamics, global temporal dependencies, and inter-vessel spatial topology.

        \subsubsection{Feature Embedding and Temporal Graph Encoding}
        
        Let $T_A = \{ \tau_1^A, \dots, \tau_N^A \}$ and $T_V = \{ \tau_1^V, \dots, \tau_M^V \}$ denote the sets of AIS and visual trajectories within a time window. Each trajectory $\tau_i$ is a sequence of unified coordinates that is projected into a high-dimensional feature space using a linear embedding layer. To capture vessel motion patterns, we construct a dynamic temporal graph for each trajectory. Its nodes represent time steps, while its edges include both adjacent temporal connections and k-nearest neighbor connections selected according to the similarity of normalized trajectory coordinates. A graph attention network (GAT) encoder \cite{velickovic2017graph} is then employed to aggregate information along these edges,

        \begin{equation}
            H_{g a t}=\operatorname{GAT}\left(\mathcal{G}_{t i m e}, X_{s e q}\right),
        \end{equation}

        \noindent where $\mathcal{G}_{time}$ is the adjacency matrix constructed from the temporal order and coordinate similarity within each trajectory. This graph allows nonadjacent time steps with similar motion states to exchange information.
        
        \subsubsection{Dual-Stream Spatio-Temporal Attention}

        Parallel to the GAT branch, we introduce a dual-stream attention mechanism to refine feature representations. To capture long-range dependencies across the entire time window, we apply self-attention along the temporal dimension of each trajectory. To model the interactions among different vessels, e.g., formation sailing or collision avoidance behaviors, we apply spatial attention across the batch of trajectories within the same modality. Given the temporally aggregated trajectory features $H_{temp} \in \mathbb{R}^{N \times d}$, the spatial interaction is computed as follows

        \begin{equation}
            {H_{\text {spatial }}=H_{temp}+\operatorname{Attention}(H_{temp},H_{temp},H_{temp}).}
        \end{equation}

        This operation enables the network to learn the relative topological structure of the vessel group and reduces its sensitivity to UAV ego-motion.

        \subsubsection{Multi-Scale Fusion and Similarity Metric Learning}
        
        The GAT output is first incorporated into the temporally aggregated representation $H_{temp}$. We then fuse $H_{temp}$ and $H_{spatial}$ using a feed-forward network (FFN) with residual connections,

        \begin{equation}
            {H_{\text {final }}=\mathbf{FFN}\left(\left[H_{\text {temp }}\|H_{\text {spatial }}\right]\right)+H_{\text {temp }}+H_{\text {spatial }}.}
        \end{equation}

        Finally, the features of AIS trajectory $i$ and visual trajectory $j$ are concatenated and passed to an uncertainty-aware scoring head. The head predicts a matching logit $\mu_{i,j}$ and an uncertainty term $\sigma_{i,j}$, which attenuates ambiguous pair scores,

        \begin{equation}
            {\begin{aligned}
            \mathbf{z}_{i,j}&=\left[H_{\text {final },i}^{A}\|H_{\text {final },j}^{V}\right],\\
            \sigma_{i,j}&=\sqrt{\exp\!\left(\operatorname{Softplus}(f_{\sigma}(\mathbf{z}_{i,j}))\right)+\epsilon},\\
            S_{i,j}&=f_{\mu}(\mathbf{z}_{i,j})\exp(-\gamma\sigma_{i,j}).
            \end{aligned}}
        \end{equation}

        The network is optimized using a class-balanced binary cross-entropy loss together with bidirectional assignment-ranking terms in the AIS-to-visual and visual-to-AIS directions. During inference, the Hungarian algorithm produces one-to-one assignments, and a threshold selected on the validation set rejects low-confidence matches.

    % \usepackage{multirow}
    \begin{table*}[]
    	\centering
    	\caption{The quantitative comparison of MAir-YOLO and baseline detectors on the 12K-image subset of the MAP-M2UAV50 dataset. The suffixes n and s denote the nano and small model variants, respectively, and OBB denotes oriented bounding box. MAir-YOLOv8s-OBB is selected as the detector in AMGT based on its accuracy–efficiency trade-off.} 
    	\begin{tabular}{cccccccc}
    		\hline
            \textbf{Detection Methods}
            & \textbf{Precision ($\%$) $\uparrow$}
            & \textbf{Recall ($\%$) $\uparrow$}
            & \textbf{mAP50 ($\%$) $\uparrow$}
            & \textbf{mAP50-95 ($\%$) $\uparrow$}
            & \textbf{Params (M) $\downarrow$}
            & \textbf{GFLOPS $\downarrow$}
            & \textbf{Time (ms) $\downarrow$}
            \\ \hline
    		YOLOv8n-OBB        & 94.62          & 84.21          & 93.04          & 77.11          & \phantom{0}3.08          & \underline{\phantom{0}8.44} & \textbf{2.37} \\
    		YOLOv8s-OBB        & 90.59          & 88.65          & 93.16          & 78.65          & 11.42         & 29.63         & \textbf{2.37} \\
    		YOLOv11n-OBB       & 94.63          & 84.29          & 93.09          & 77.08          & \textbf{\phantom{0}2.66} & \textbf{\phantom{0}6.69} & 2.91 \\
    		YOLOv11s-OBB       & 93.65          & 86.55          & 93.11          & 78.54          & \phantom{0}9.71          & 22.53         & 3.12 \\          
    		MAir-YOLOv8n-OBB  & 96.58          & \underline{94.00} & 97.77          & 80.81          & \phantom{0}2.92          & 12.10         & 2.68 \\
    		MAir-YOLOv8s-OBB  & \textbf{97.23} & 93.55          & \underline{97.81} & \textbf{82.00} & 10.62         & 35.93         & \underline{2.58} \\
    		MAir-YOLOv11n-OBB & 96.57          & 93.56          & 97.55          & 80.12          & \underline{\phantom{0}2.74} & 10.96         & 3.38 \\
    		MAir-YOLOv11s-OBB & \underline{96.74} & \textbf{94.17} & \textbf{97.85} & \underline{81.75} & \phantom{0}9.88          & 31.09         & 3.51 \\ \hline               
    	\end{tabular}
    	\label{tab: yolov8-obb}
    \end{table*}

\section{Experiments and Discussion}
\label{5}

    \subsection{Implementation Details}

        \subsubsection{Experimental Setup}
            Our multi-rotor unmanned flight platform is built to collect real-world multi-sensor data and validate the effectiveness of the proposed method. The devices equipped on the UAV mainly include a camera, GNSS, IMU, AIS receiver, and a computing device. More details can be found in \cite{10833660}. The multi-sensor data are collected by the dynamic UAV-based platform from the Wuhan segment of the Yangtze River. The deep-learning-based models are trained and tested on a PC with Intel Core i9-14900KF CPU @ 3.2 GHz and Nvidia RTX 4090D GPU. For the graph learning-based association method, the Adam optimizer is used during training, with a learning rate of 0.0001 to ensure stable and efficient parameter updates. The model is trained for 100 epochs with a window size setting as 15 seconds.
                 
        \subsubsection{Evaluation Metrics}
            We evaluate the proposed framework from three aspects, i.e., vessel detection, multi-object tracking and association. For vessel detection, Precision and Recall measure the correctness and completeness of detected objects, respectively. The mean Average Precision (mAP) is adopted as the primary detection metric, where mAP50 denotes the mAP at an IoU threshold of 0.5, and mAP50-95 denotes the average mAP over IoU thresholds from 0.5 to 0.95 with a step of 0.05. The number of parameters (Params), GFLOPS, and inference time are further reported to evaluate computational complexity and real-time efficiency.
            
            For multi-object tracking, we report identity switches (IDS), multiple object tracking accuracy (MOTA), multiple object tracking precision (MOTP), and identity F1 score (IDF1). IDS counts changes in the predicted identity assigned to a ground-truth target. MOTA accounts for false positives, false negatives, and identity switches, and MOTP measures localization accuracy for matched targets \cite{bernardin2008clear}. IDF1 evaluates identity consistency over trajectories \cite{ristani2016performance}. We additionally report HOTA, which jointly evaluates detection and association performance \cite{luiten2021hota}.
            
            For visual-AIS association, association precision, association recall, association F1, and association ID switches (A-IDS) are reported. A-IDS counts changes in the predicted AIS identity for the same visual track across consecutive evaluated windows. Lower A-IDS indicates more stable association identities. The first three metrics are defined as follows

            \begin{equation}
                {
                \begin{gathered}
                \operatorname{Precision}
                = N_{\mathrm{correct}} / N_{\mathrm{pred}},\\[3pt]
                \operatorname{Recall}
                = N_{\mathrm{correct}} / N_{\mathrm{gt}},\\[3pt]
                \operatorname{F1}
                = 2\operatorname{Precision}\cdot\operatorname{Recall}
                  / \left(\operatorname{Precision}+\operatorname{Recall}\right).
                \end{gathered}
                }
            \label{eq:association_metrics}
            \end{equation}

            \noindent where $N_{correct}$, $N_{pred}$, and $N_{gt}$ denote the numbers of correct predicted pairs, all predicted pairs, and ground-truth pairs, respectively.

    \subsection{Experiments on Maritime Object Detection}
    
    Comparative experiments were conducted using the YOLOv8-OBB and YOLOv11-OBB series. As shown in Table \ref{tab: yolov8-obb}, MAir-YOLOv8s-OBB achieves an mAP50 of 97.81$\%$ and an mAP50-95 of 82.00$\%$. Compared with the corresponding YOLOv8s-OBB baseline, these represent gains of 4.65 and 3.35 percentage points, respectively. Its parameter count decreases by 0.80 M, while its computational cost increases by 6.30 GFLOPS. The measured FP32 model-forward time is 2.58 ms per image on the RTX 4090D under the protocol specified in Table \ref{tab: yolov8-obb}. MAir-YOLOv8s-OBB is therefore selected as the detector in AMGT because it achieves the highest precision and mAP50-95 among the evaluated models, despite its higher computational cost.

    \subsection{Experiments on Multi-Vessel Tracking}
        % \usepackage{multirow} 
        \begin{table*}[]
        	\centering
        	\caption{The overall tracking performance on eight representative airborne maritime video sequences using MAir-YOLOv8s-OBB detection results. Time denotes the average tracking update time per frame.} 
        	\begin{tabular}{ccccccccc}
            \hline
            \textbf{Tracking Methods}
            & \textbf{IDS} $\downarrow$
            & \textbf{Precision ($\%$) $\uparrow$}
            & \textbf{Recall ($\%$) $\uparrow$}
            & \textbf{MOTA ($\%$) $\uparrow$}
            & \textbf{MOTP ($\%$) $\uparrow$}
            & \textbf{IDF1 ($\%$) $\uparrow$}
            & \textbf{HOTA ($\%$) $\uparrow$}
            & \textbf{Time (ms) $\downarrow$}
            \\ \hline
            SORT \cite{7533003} \cite{10833660}                     & 56            & 92.31                    & 89.94                 & 81.97               & 78.28               & 79.24               & 65.91               & \textbf{0.182}            \\
            ByteTrack \cite{zhang2022bytetrack}                & 63            & 89.46                    & 90.05                 & 78.90               & 77.67               & 78.87               & 64.87               & 0.718                     \\
            BoT-SORT \cite{aharon2022bot}                  & 12            & 90.05                    & \textbf{92.31}        & 82.00               & 79.27               & 89.51               & 71.36               & 10.801                    \\
            MAir-SORT                 & \textbf{2}    & \textbf{92.80}           & 90.85                 & \textbf{83.78}      & \textbf{79.53}      & \textbf{91.76}      & \textbf{72.77}      & 2.874                     \\ \hline
            \end{tabular}
        	\label{tab: tracking}
        \end{table*}
        \begin{figure*}
        	\centering
        	\includegraphics[width=1 \textwidth]{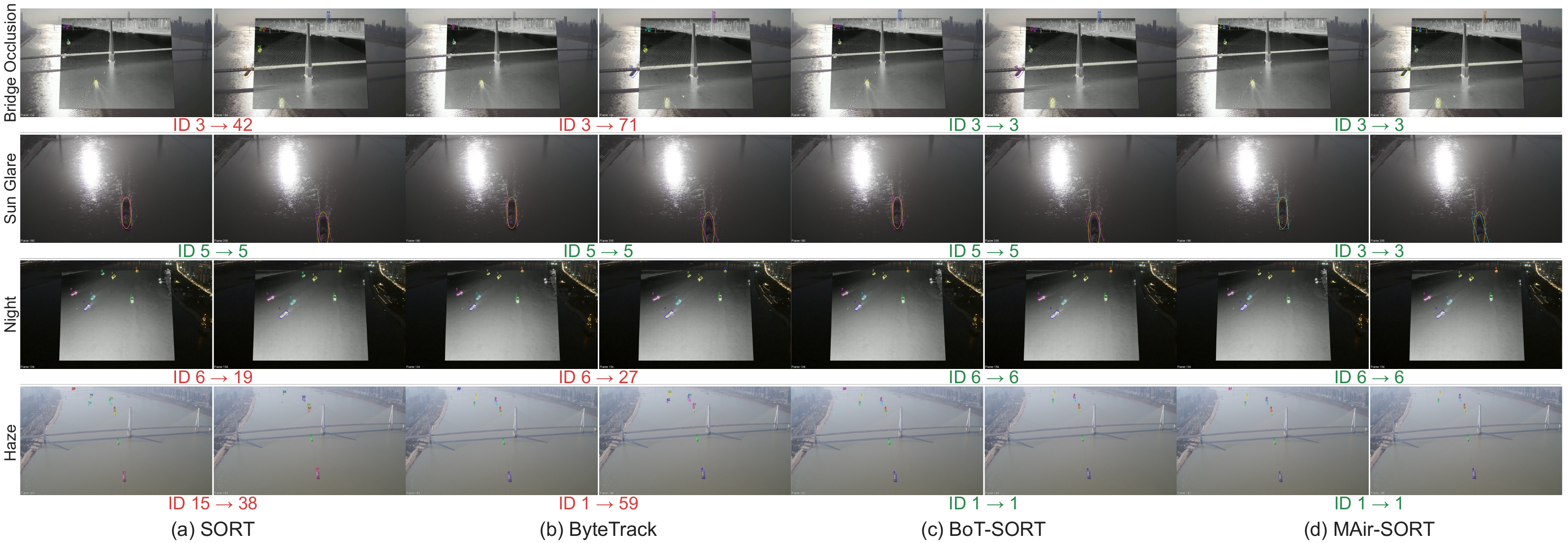}
        	\caption{The qualitative comparison of SORT \cite{7533003} \cite{10833660}, ByteTrack \cite{zhang2022bytetrack}, BoT-SORT \cite{aharon2022bot}, and MAir-SORT under bridge occlusion, sun glare, nighttime imaging, and haze. Columns correspond to tracking methods, and rows correspond to scene conditions. Each method is shown at two frames from the same sequence. The ID pair below the images gives the identities assigned to the same selected vessel at the two frames. Green text indicates an unchanged identity, while red text indicates an identity switch. The tracked IDs are generated independently by each method.} 
        	\label{fig: tracking} 
        	\vspace{-1.0em}
        \end{figure*}

        We evaluate the four tracking methods on eight representative airborne maritime sequences under sun glare, bridge occlusion, haze, nighttime imaging conditions. All methods use detections produced by the same MAir-YOLOv8s-OBB model, while the ground-truth boxes are used only for evaluation. Table \ref{tab: tracking} shows that MAir-SORT produces only 2 identity switches, compared with 56 for SORT, 63 for ByteTrack, and 12 for BoT-SORT. It also achieves the highest precision of 92.80\%, MOTA of 83.78\%, MOTP of 79.53\%, IDF1 of 91.76\%, and HOTA of 72.77\%. Compared with SORT, MAir-SORT improves MOTA, IDF1, and HOTA by 1.81, 12.52, and 6.86 percentage points, respectively. Its much larger gains in IDF1 and HOTA than in MOTP indicate that the main improvement lies in identity continuity rather than box localization, which is consistent with the intended roles of camera-motion compensation, multi-cue association, and lost-track recovery. BoT-SORT provides the highest recall of 92.31\%, while SORT has the shortest update time of 0.182 ms per frame. MAir-SORT requires 2.874 ms per frame, providing a practical balance between identity preservation and tracking efficiency under the evaluated protocol.
    
        The qualitative comparison of the four trackers under challenging conditions is presented in Fig. \ref{fig: tracking}. Each row shows two frames from one scene, and the ID pair below each method records the identities assigned to the same selected vessel. Since track IDs are generated independently, their absolute values should not be compared across methods. In the bridge-occlusion, nighttime, and hazy conditions, SORT and ByteTrack assign new identities, whereas BoT-SORT and MAir-SORT maintain identity continuity. All four methods retain the selected identity under sun glare because the low traffic density makes this scene relatively simple. Overall, these qualitative results illustrate the stronger identity continuity and robustness of MAir-SORT under challenging maritime conditions.

    \subsection{Experiments on Multi-Vessel Association}
\begin{figure}
        \centering
        \includegraphics[width=0.50 \textwidth]{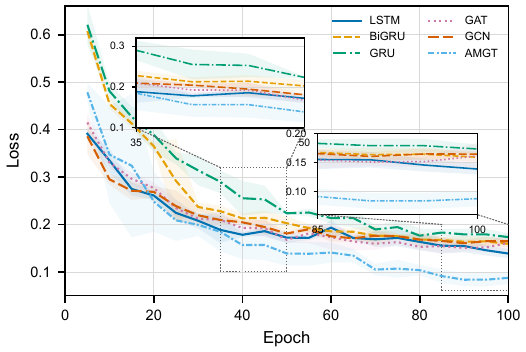}
        \caption{The training loss convergence of the six learning-based association methods over 100 epochs. Curves and shaded regions denote the mean and sample standard deviation over three random seeds. The inset panels enlarge epochs 35-50 and 85-100.}
        \label{fig: association loss} 
        \vspace{-1.0em}
        \end{figure}

        To validate the robustness of the proposed AMGT framework, we conducted a comprehensive comparison against three traditional geometric methods, i.e., Chebyshev distance (CD) \cite{lin2023infocd}, dynamic time warping (DTW) \cite{9970291}, and Euclidean distance (ED) \cite{9965921}, and five data-driven learning baselines. The learning-based methods are evaluated over three random seeds under the same data partition. Fig. \ref{fig: association loss} shows that all six models decrease their training losses, while AMGT reaches the lowest final loss with a relatively narrow variation band. Since the loss scales depend on the model architecture, the curves are used to assess optimization convergence rather than association accuracy.

        \begin{table*}[t]
        \centering
        \caption{The quantitative visual-AIS association comparison with and without motion-aware coordinate unification (MCU). All methods use the same MAir-YOLOv8s-OBB detections and MAir-SORT outputs. Learning-based methods report the mean $\pm$ sample standard deviation over three random seeds.}
        \begin{tabular}{ccccccc}
        \hline
        \textbf{Input}
        & \textbf{Association Methods}
        & \textbf{Precision ($\%$) $\uparrow$}
        & \textbf{Recall ($\%$) $\uparrow$}
        & \textbf{F1 ($\%$) $\uparrow$}
        & \textbf{A-IDS} $\downarrow$
        & \textbf{Time (ms/window) $\downarrow$}
        \\ \hline
                                      & CD \cite{lin2023infocd}                                        & 14.49                                   & 11.81                                & 13.02                            & 17                              & 0.198                                     \\
                                      & DTW \cite{9970291}                                       & 15.94                                   & 12.99                                & 14.32                            & 17                              & 5.61                                      \\
                                      & ED \cite{9965921}                                        & 15.94                                   & 12.99                                & 14.32                            & 17                              & \textbf{0.081}                                     \\
                                      & LSTM \cite{graves2012long}                                      & 84.36 ± 3.09                            & 82.81 ± 2.85                         & 83.58 ± 2.97                     & 9.00 ± 1.00                     & 0.721 ± 0.003                             \\
                                      & BiGRU \cite{cho2014learning}                                     & 93.69 ± 1.50                            & 89.24 ± 4.21                         & 91.37 ± 2.14                     & 8.67 ± 0.58                     & 0.514 ± 0.001                             \\
                                      & GRU \cite{chung2014empirical}                                       & 91.73 ± 2.27                            & 90.29 ± 2.24                         & 91.01 ± 2.26                     & 13.00 ± 3.46                    & 0.514 ± 0.001                             \\
                                      & GAT \cite{velickovic2017graph}                                       & 93.71 ± 0.46                            & 91.86 ± 0.45                         & 92.78 ± 0.46                     & 8.67 ± 1.15                     & 1.333 ± 0.023                             \\
                                      & GCN \cite{kipf2016semi}                                       & 85.11 ± 13.04                           & 83.07 ± 13.41                        & 84.08 ± 13.23                    & 12.00 ± 6.08                    & 3.621 ± 0.009                             \\
        \multirow{-9}{*}{Without MCU} & Our AMGT                                      & \textbf{98.53 ± 1.15}                            & \textbf{96.98 ± 1.14}                         & \textbf{97.75 ± 1.15}                     & \textbf{1.33 ± 2.31}                     & 3.807 ± 0.002                             \\ \hline
                                      & CD \cite{lin2023infocd}                                        & 93.2                                    & 91.73                                & 92.46                            & 9                               & 0.189                                     \\
                                      & DTW \cite{9970291}                                       & 92.4                                    & 90.94                                & 91.67                            & 8                               & 5.339                                     \\
                                      & ED \cite{9965921}                                        & 92.4                                    & 90.94                                & 91.67                            & 8                               & \textbf{0.078}                                     \\
                                      & LSTM \cite{graves2012long}                                      & 96.26 ± 1.66                            & 94.49 ± 1.36                         & 95.37 ± 1.50                     & 5.00 ± 2.00                     & 0.727 ± 0.043                             \\
                                      & BiGRU \cite{cho2014learning}                                     & 94.40 ± 1.83                            & 92.91 ± 1.80                         & 93.65 ± 1.82                     & 7.00 ± 3.00                     & 0.514 ± 0.001                             \\
                                      & GRU \cite{chung2014empirical}                                       & 95.20 ± 1.06                            & 93.70 ± 1.04                         & 94.44 ± 1.05                     & 5.67 ± 0.58                     & 0.513 ± 0.000                             \\
                                      & GAT \cite{velickovic2017graph}                                       & 97.87 ± 0.61                            & 96.33 ± 0.60                         & 97.09 ± 0.61                     & 3.00 ± 1.73                     & 1.333 ± 0.023                             \\
                                      & GCN \cite{kipf2016semi}                                       & 96.80 ± 1.60                            & 95.28 ± 1.57                         & 96.03 ± 1.59                     & 4.00 ± 1.00                     & 3.575 ± 0.008                             \\
        \multirow{-9}{*}{With MCU}    & Our AMGT                                      & \textbf{98.40 ± 0.40}                            & \textbf{96.85 ± 0.39}                         & \textbf{97.62 ± 0.40}                     & \textbf{2.33 ± 2.52}                     & 3.803 ± 0.014                             \\ \hline
        \end{tabular}
        \label{tab: association}
        \end{table*}

        Table \ref{tab: association} reports the results with and without MCU. AMGT without MCU obtains the highest precision, recall, and F1 of 98.53\%, 96.98\%, and 97.75\%, respectively, with 1.33 association identity switches. With MCU, it retains the best precision, recall, and F1 of 98.40\%, 96.85\%, and 97.62\%, respectively. MCU substantially improves the geometric methods and several learning baselines, showing that coordinate alignment benefits methods that depend strongly on absolute trajectory positions. The small AMGT difference between the two inputs is within the observed three-seed variation and therefore indicates robustness to the coordinate representation rather than a measurable gain from MCU. ED has the lowest pure inference time, while the sequential dynamic programming in DTW leads to a higher latency.

        \begin{figure*}
        \centering
        \includegraphics[width=1 \textwidth]{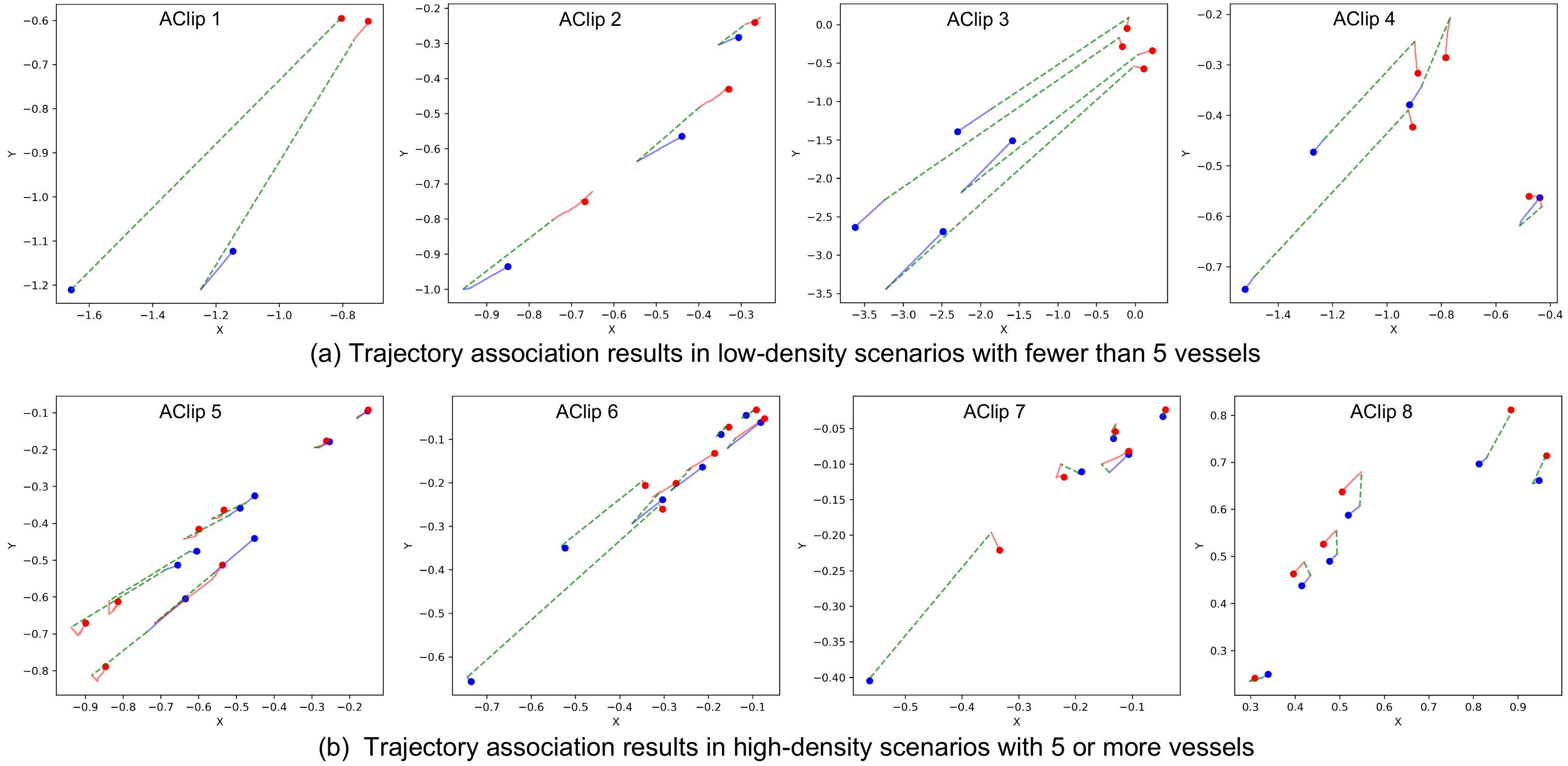}
        \caption{The visualization of the trajectory association under varying traffic conditions. The maritime scenarios are broadly classified into: (a) Low-density and (b) High-density. The green dashed lines (predicted associations) accurately connect the visual trajectories (red lines) with their corresponding AIS trajectories (blue lines). The low-density scenarios have clear spatial layouts, while high-density scenarios involve intricate inter-vessel interactions and overlaps.}
        \label{fig: association results} 
        \vspace{-1.0em}
        \end{figure*}
        \begin{figure*}
        \centering
        \includegraphics[width=1 \textwidth]{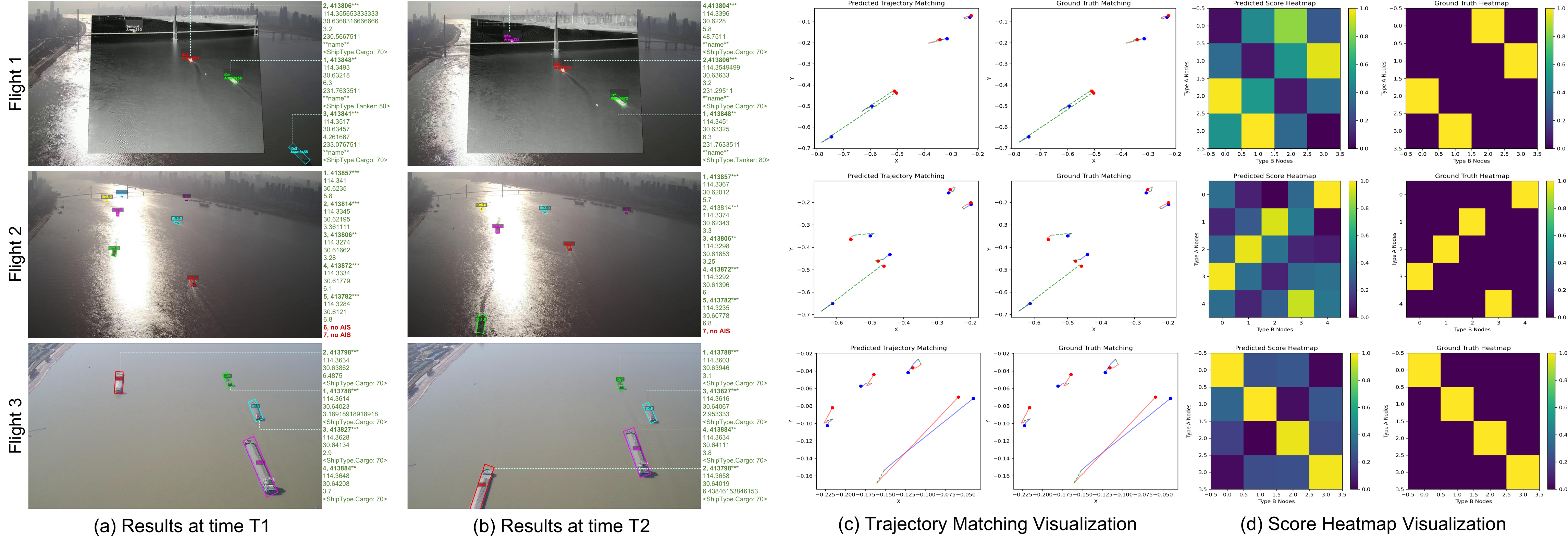}
        \caption{The visualization of multi-vessel identification and tracking under three UAV flight motions (From top to bottom: Flight 1, 2, and 3, which are shown in Fig. \ref{fig: coor_unification}. From left to right: (a) and (b) Results at time T1 and T2, (c) Trajectory matching by prediction (left) and ground truth (GT, right), (d) Score heatmap by prediction (left) and GT (right).}
        \label{fig: VIT results} 
        \vspace{-1.0em}
        \end{figure*}
        Fig. \ref{fig: association results} provides a comprehensive visualization of the trajectory association results across various distinct clips, including AClip 1–8. They are categorized by vessel density into low-density ($<5$ vessels) and high-density ($\ge5$ vessels) groups. In low-density scenarios, the topological structure is relatively simple between surface vessels. Notably, even when significant spatial offsets exist between the two modalities due to position errors, our method correctly identifies the matches, demonstrating that it does not rely solely on absolute position overlap. Fig. \ref{fig: association results}(b) presents much more challenging maritime environments. In AClip 5 and AClip 6, vessels move in tight formations, resulting in highly overlapping trajectories and intricate spatial interactions. Our AMGT framework successfully resolves these ambiguities. By modeling the relative topological graph, the network discerns the correct identity even when individual positions are visually crowded.
        We further validate the robustness of the proposed AMGT framework across three distinct UAV flight dynamics, i.e. rapid motion (Flight 1), hovering (Flight 2), and slow motion (Flight 3), as defined in Fig. \ref{fig: coor_unification}. The qualitative results are comprehensively illustrated in Fig. \ref{fig: VIT results}. The first two columns display the real-time identification and tracking results at different timestamps ($T_1$ and $T_2$). The associated AIS information is overlaid on the right side of the image, with sensitive details (e.g., MMSI and vessel name) anonymized. The third column visualizes the trajectory matching on a unified normalized plane. Despite the UAV's motion, the visual trajectories (red) and AIS trajectories (blue) exhibit high spatial consistency, verifying the effectiveness of the proposed coordinate unification method. The final column presents the predicted similarity heatmaps. The clear diagonal patterns indicate high-confidence matching. Notably, in Flight 2, vessels with tracking IDs 7 and 8 are identified to lack corresponding AIS data. The AMGT framework correctly handles them as unassociated visual objects without forcing erroneous matches, demonstrating its capability to distinguish vessels without AIS information.

   \subsection{Discussion}
        The experiments show that the three main stages of AMGT address complementary parts of the UAV-based maritime perception pipeline. MAir-YOLOv8s-OBB provides oriented vessel detections under different imaging conditions, while MAir-SORT reduces identity switches relative to the compared trackers. For visual-AIS association, AMGT achieves the highest F1 under both coordinate inputs, supporting the use of temporal graph encoding and inter-vessel attention for heterogeneous trajectory matching.

        The coordinate unification clearly improves the geometric methods and several learning baselines, but it does not increase the mean F1 in the current three-seed evaluation. This suggests that AMGT is comparatively robust to the tested coordinate representations, while the benefit of explicit motion compensation remains method dependent. The evaluation is limited to eight association sequences, and broader validation under more diverse UAV motion, vessel density, and AIS availability is still needed.

\section{Conclusion}
\label{6}
This paper proposed a multi-modal graph learning-driven framework, termed AMGT, for vessel identification and tracking in UAV-assisted maritime IoT. AMGT addresses sensor heterogeneity, adverse imaging conditions, and spatial misalignment caused by UAV motion. Visible and infrared images are fused to enhance visual perception, while AIS messages are temporally synchronized to provide vessel identity and motion information. A mixed-modality MAir-YOLOv8-OBB detector and MAir-SORT tracker are developed for oriented vessel detection and stable trajectory tracking. In addition, a motion-aware coordinate unification algorithm reduces nonlinear spatial distortions, and a spatio-temporal graph neural network associates visual observations with AIS data by exploiting relative structural consistency. Experiments on the self-collected maritime dataset show that AMGT outperforms existing methods. Its computational efficiency also indicates potential for edge deployment in UAV-assisted maritime surveillance.

Future work will focus on three directions. First, more advanced visible–infrared image registration and fusion methods will be investigated to improve cross-modal alignment and exploit complementary spectral information under challenging maritime conditions. In particular, both robust registration strategies and learning-based fusion approaches will be explored to alleviate spatial misalignment and enhance the quality of multimodal visual perception. Second, global motion smoothing and long-term temporal attention will be explored to improve tracking stability under aggressive UAV maneuvers. Third, AMGT will be extended to distributed multi-UAV surveillance. Inspired by recent studies on air–ground collaborative networking and UAV-assisted IoT data collection \cite{11270860}, we will investigate edge-enabled collaborative sensing among multiple UAVs. Local topological graphs will be shared or compressed at the edge to fuse cross-view information, reduce occlusion effects, and expand surveillance coverage.

% \section*{Acknowledgments}
% This should be a simple paragraph before the References to thank those individuals and institutions who have supported your work on this article.



\bibliographystyle{IEEEtran}
\bibliography{reference}

\newpage

\vfill

\end{document}